\documentclass{article} 
\usepackage[final]{colm2026_conference}

\usepackage[T1]{fontenc}
\usepackage[utf8]{inputenc} 

\usepackage{microtype}
\usepackage{hyperref}
\usepackage{url}
\usepackage{booktabs}

\usepackage{lineno}

\definecolor{darkblue}{rgb}{0, 0, 0.5}
\hypersetup{colorlinks=true, citecolor=darkblue, linkcolor=darkblue, urlcolor=darkblue}

\usepackage{multirow}
\usepackage{float}
\usepackage[most]{tcolorbox}
\usepackage[dvipsnames]{xcolor}
\definecolor{DarkGreen}{RGB}{0,100,0}
\newtcolorbox{insightbox}{%
  colback=blue!6,
  colframe=blue!40!black,
  boxrule=0.8pt,
  arc=3pt,
  left=6pt,
  right=6pt,
  top=3pt,
  bottom=3pt
}

\title{Data Repetition Beats Data Scaling in Long-CoT Supervised Fine-Tuning}

\author{
  \textbf{Dawid J. Kopiczko}\textsuperscript{1}\quad
  \textbf{Sagar Vaze}\textsuperscript{2}\quad
  \textbf{Tijmen Blankevoort}\textsuperscript{3}\quad
  \textbf{Yuki M. Asano}\textsuperscript{1}
  \\[2pt]
  \normalfont
  \textsuperscript{1}University of Technology Nuremberg
  \quad
  \textsuperscript{2}Mistral AI
  \quad
  \textsuperscript{3}NVIDIA
  \\[2pt]
  \texttt{dj.kopiczko@gmail.com}
}

\begin{document}

\ifcolmsubmission
\linenumbers
\fi

\maketitle

\begin{abstract}
Supervised fine-tuning (SFT) on chain-of-thought data is an essential post-training step for reasoning language models.
Standard machine learning intuition suggests that training with more unique training samples yields better generalization. Counterintuitively, we show that SFT benefits from \emph{repetition}: under a fixed update budget, training for more epochs on smaller datasets outperforms single-epoch training on larger datasets. On AIME'24/25 and GPQA benchmarks, Olmo3-7B trained for 128 epochs on 400 long chain-of-thought samples outperforms the equivalent 1 epoch on 51200 samples by 12--26 percentage points, with no additional catastrophic forgetting.
We find that training token accuracy reliably signals when repetition has saturated, as improvements from additional epochs plateau at full memorization, a pattern consistent across all settings. These findings provide a practical
approach for reasoning SFT, where scaling epochs with token accuracy as a stopping criterion can replace expensive undirected data scaling.
We pose the \emph{repetition advantage}, where full memorization coincides with improved generalization, as a new open problem for the community in understanding the training dynamics of large language models.
\end{abstract}

\section{Introduction}

Modern language model training proceeds through distinct stages: pretraining on internet-scale data to acquire world knowledge, mid-training on curated corpora to extend capabilities, and post-training to shape model behavior \citep{guo2025deepseekr1,teamolmo2025olmo3,yang2025qwen3}. For reasoning-focused models, post-training typically begins with supervised fine-tuning (SFT) on long Chain-of-Thought (CoT) demonstrations, often distilled from more capable models, where reasoning traces can span thousands of tokens before reaching a final answer. 
This SFT step, analogous to behavioral cloning in reinforcement learning \citep{osa2018imitation}, primes the model for subsequent stages such as reinforcement learning from human feedback \citep{ouyang2022instructgpt} or reinforcement learning with verifiable rewards \citep{guo2025deepseekr1, deepseek-math}. 
Unlike pretraining data, which can be scraped at scale from the web, high-quality long-CoT demonstrations require either expensive human annotation or careful distillation from larger models, including generation, filtering, and validation of long reasoning traces. As a result, the question of how to best utilize limited SFT data is practically important.

\begin{figure*}[t]
  \begin{center}
    \centerline{\includegraphics[width=\textwidth]{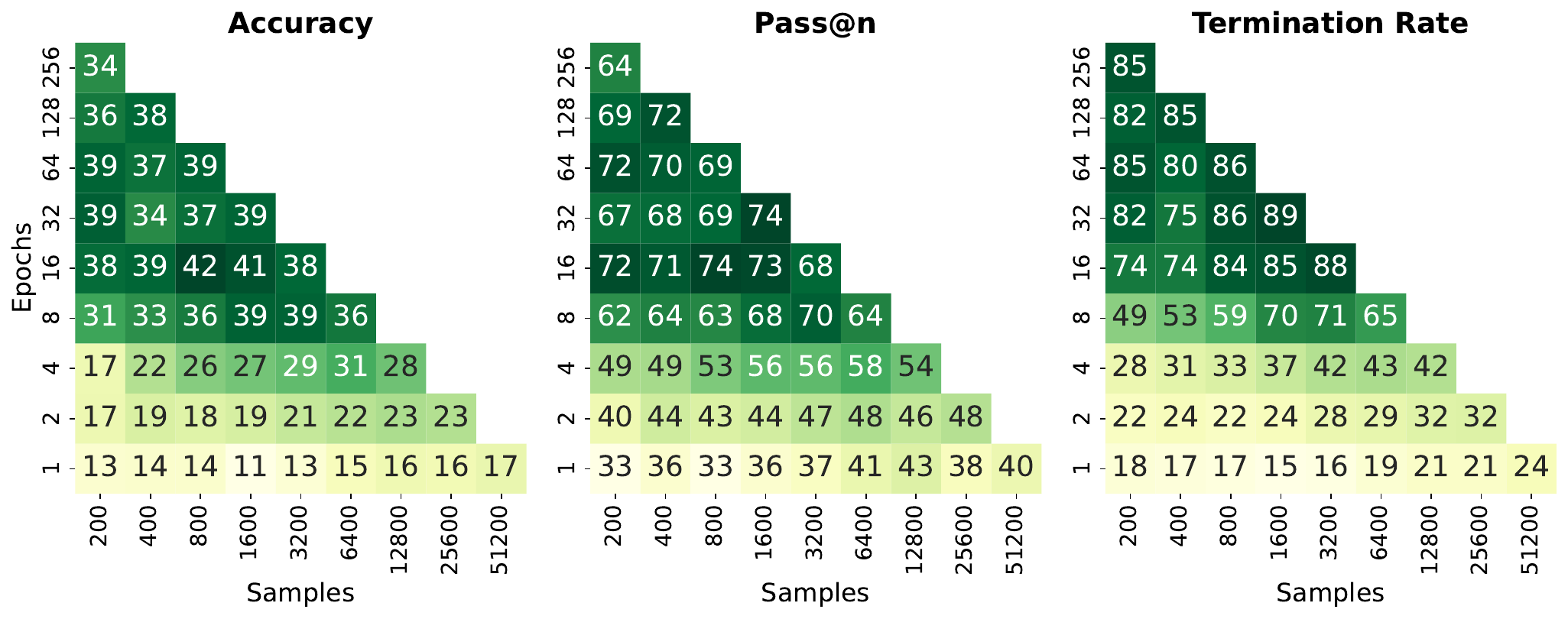}}
    \caption{
    Scaling epochs versus scaling data for Olmo3-7B trained on long-CoT SFT data, averaged across AIME'24, AIME'25, and GPQA benchmarks. Each diagonal represents a fixed update budget, where epochs × samples is constant. Within any diagonal, moving toward fewer samples and more epochs consistently improves accuracy and pass@n, with gains diminishing around 32–64 epochs. Termination rate correlates strongly with accuracy.
    }
    \label{fig1}
  \end{center}
\end{figure*}

The common assumption in machine learning would suggest that training with more unique training samples yields better generalization. Under i.i.d.\ sampling, each new example provides independent information about the data distribution, and generalization bounds in statistical learning theory typically improve with dataset size. This principle manifests practically throughout the field -- data augmentation techniques are widely used to artificially expand effective dataset size when real data is limited~\citep{dataAugment, shortenSurveyImageData2019}, and the success of large language models has been attributed in significant part to training on ever-larger unique corpora. Following this logic, modern post-training pipelines employ millions of SFT samples~\citep{teamolmo2025olmo3}.

In this paper, we show that this might not only be suboptimal, but that, actually, a reverse pattern can be observed for the SFT stage for a pretrained LLM. 
Under a fixed update budget, \textbf{training for more epochs on smaller datasets outperforms training on
larger datasets}. The gains are not marginal. The performance and termination rates, i.e., the model's ability to successfully conclude reasoning with a final answer, both scale with epoch count and saturate together, suggesting that sufficient repetition of the same data is required for models to fully internalize the demonstrated reasoning structure.

We find that this convergence is tightly linked to \textbf{training set memorization}. Performance improvements plateau once models achieve near-perfect next-token prediction accuracy on the training data, even as validation loss continues to rise. This relationship holds across all models we test, all benchmarks, and different training datasets, making train token accuracy a practical stopping criterion for scaling epochs. Despite this apparent overfitting, we observe no additional catastrophic forgetting compared to single-epoch training on large datasets.
Our main contributions are:
\begin{itemize}
\itemsep0em
    \item \textbf{Phenomenon}. We demonstrate that under a fixed update budget, scaling epochs on smaller datasets substantially outperforms scaling unique samples.
    \item \textbf{Dynamics}. We identify training token accuracy as a reliable stopping criterion for epoch scaling, with performance gains plateauing once models reach full memorization, and we show that multi-epoch training on small datasets causes no additional catastrophic forgetting compared to
    large datasets.
    \item \textbf{Factors}. We show how training data properties, such as teacher model size in distillation and the correctness of data samples, affect the repetition advantage.
\end{itemize}
While we provide a practical heuristic for exploiting the repetition advantage, we pose explaining this phenomenon in long-CoT SFT as a novel, open problem for the community.

\begin{figure*}[t]
  \vskip -0.2in
  \begin{center}
    \centerline{\includegraphics[width=\textwidth]{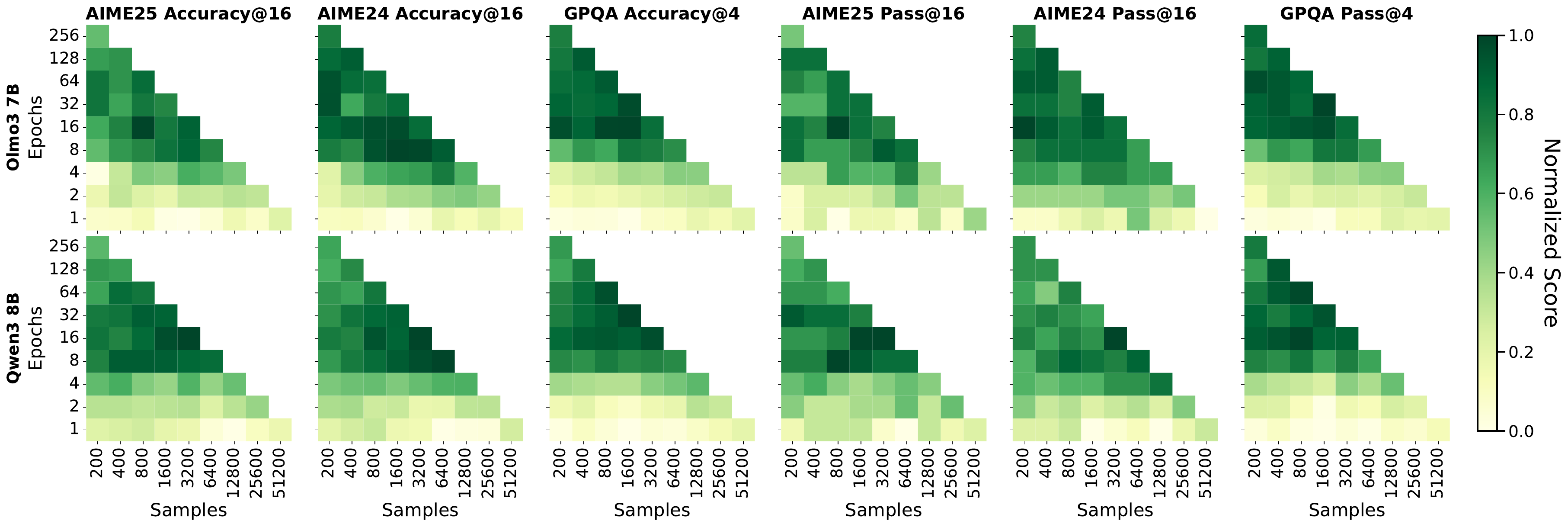}}
    \caption{
    The repetition advantage is consistent across models, benchmarks, and evaluation metrics. Heatmaps show normalized scores for Olmo3-7B (top) and Qwen3-8B (bottom) on AIME'24, AIME'25, and GPQA. Each diagonal corresponds to a fixed update budget (epochs $\times$ samples), and in all settings, performance improves when moving along a diagonal toward fewer samples and more epochs.
    }
    \label{fig_heatmaps}
  \end{center}
\end{figure*}

\section{Scaling Epochs on a Fixed Update Budget}
\label{sec:main}

To investigate whether data repetition can substitute for data scaling in supervised fine-tuning, we conduct controlled experiments varying the number of epochs and unique samples while holding total gradient updates, and all other parameters, constant. We train base checkpoints of two recent language models on chain-of-thought data and evaluate on challenging reasoning benchmarks.

\subsection{Preliminaries}

Supervised fine-tuning adapts a pretrained language model to target behaviors by training on demonstration data. Given input-output pairs $(x, y)$ where $y = (y_1, \ldots, y_T)$ is a target sequence, SFT minimizes the cross-entropy loss over next-token predictions:
\begin{equation}
\mathcal{L}(\theta) = -\sum_{t=1}^{T} \log p_\theta(y_t \mid x, y_{<t})
\end{equation}
In practice, the loss is typically masked to exclude input tokens, applying only to the response.

Throughout this work, we use \emph{update budget} $\mathcal{B}$ to denote the total number of gradient updates during training, which for batch size one is equal to the number of epochs multiplied by the number of unique samples. Comparing configurations at equal update budgets isolates the effect of data repetition from differences in total optimization steps.

\subsection{Experimental Setup}

\noindent\textbf{Models.} We use the Qwen3-4B, Qwen3-8B \citep{yang2025qwen3}, and Olmo3-7B \citep{teamolmo2025olmo3} base models. These are pretrained checkpoints prior to any instruction tuning, providing a clean starting point for studying SFT dynamics. For training and evaluation, we use the default chat template for each model.

\noindent\textbf{Dataset.} We use the \emph{Dolci SFT 7B}\footnote{\url{https://huggingface.co/datasets/allenai/Dolci-Think-SFT-7B}} dataset from the Olmo3 post-training pipeline, which contains distilled long-CoT demonstrations spanning math, coding, precise instruction following, and general conversation \citep{teamolmo2025olmo3}. We apply several filters: we keep only the first conversation turn, retain samples containing complete reasoning traces (verified by presence of \texttt{<think>} and \texttt{</think>} tags), and remove samples exceeding 10k tokens when tokenized with the Olmo tokenizer. From the filtered data, we randomly sample nested training splits of increasing size: 200, 400, 800, 
1.6k, 3.2k, 6.4k, 12.8k, 25.6k, 51.2k
samples, constructed so that each smaller split is a subset of the next larger one. We hold out 1000 random samples as a validation set for analysis.

\noindent\textbf{Evaluation.} We evaluate on three challenging reasoning benchmarks: AIME 2024, AIME 2025, and GPQA. AIME \citep{aops_aime_problems_solutions_2025} is a mathematical reasoning benchmark consisting of 30 competition problems per year, requiring multi-step reasoning across algebra, geometry, number theory, and combinatorics; each answer is an integer from 0 to 999. GPQA \citep{rein2024gpqa} is a graduate-level multiple-choice benchmark with expert-written questions in biology, physics, and chemistry, where the model must reason through the problem before selecting from four options. For each problem in these benchmarks, we append an instruction requesting the final answer in \texttt{\textbackslash boxed\{\}} format for straightforward extraction.

\begin{table*}[t]
\vskip -0.2in
\begin{center}
\small
\begin{tabular}{c@{\hskip 0.3cm}r@{\hskip 0.5cm}r@{\hskip 0.5cm}rr@{\hskip 0.5cm}rr@{\hskip 0.5cm}rr}
\toprule
Model & Epochs & Samples & \multicolumn{2}{c}{GPQA} & \multicolumn{2}{c}{AIME'24} & \multicolumn{2}{c}{AIME'25} \\
\cmidrule(lr){4-5}
\cmidrule(lr){6-7}
\cmidrule(lr){8-9}
 & &  & Avg@4 & Pass@4 & Avg@16 & Pass@16 & Avg@16 & Pass@16 \\
\midrule
 \multirow{6}{*}{\rotatebox{90}{Olmo3-7B}}  & 1 & 51.2k & 11.5 & 23.7 & 17.7 & 46.7 & 22.3 & 50.0 \\
 & 2 & 25.6k & 14.8 & 29.3 & 28.1 & 66.7 & 24.8 & 46.7 \\
 & 4 & 12.8k & 20.2 & 38.9 & 33.3 & \underline{73.3} & 29.2 & 50.0 \\
 & 8 & 6.4k & \underline{29.7} & \underline{51.5} & \textbf{44.4} & \underline{73.3} & \underline{35.4} & \textbf{66.7} \\
 & 16 & 3.2k & \textbf{34.0} & \textbf{62.1} & \underline{42.3} & \textbf{80.0} & \textbf{39.2} & \underline{63.3} \\
\midrule
\multirow{5}{*}{\rotatebox{90}{Qwen3-8B}} & 1 & 51.2k & 21.6 & 38.9 & 12.3 & 36.7 & 11.2 & 30.0 \\
 & 2 & 25.6k & 25.4 & 42.4 & 14.0 & 46.7 & 17.3 & 43.3 \\
 & 4 & 12.8k & 35.6 & 56.6 & \underline{20.6} & 66.7 & 20.0 & 40.0 \\
 & 8 & 6.4k & \underline{41.9} & \underline{61.6} & \textbf{30.6} & \underline{70.0} & \underline{27.7} & \underline{56.7} \\
 & 16 & 3.2k & \textbf{51.0} & \textbf{72.7} & \textbf{30.6} & \textbf{76.7} & \textbf{31.2} & \textbf{63.3} \\
\midrule
\multirow{5}{*}{\rotatebox{90}{Qwen3-4B}} & 1 & 51.2k & 13.1 & 29.8 & 6.5 & 26.7 & 5.6 & 30.0 \\
 & 2 & 25.6k & 21.1 & 38.9 & 13.5 & \underline{36.7} & 14.0 & 36.7 \\
 & 4 & 12.8k & 29.7 & 52.5 & 18.1 & \underline{36.7} & 17.7 & \underline{40.0} \\
 & 8 & 6.4k & \textbf{40.5} & \underline{64.1} & \textbf{23.3} & \textbf{43.3} & \textbf{23.5} & \textbf{43.3} \\
 & 16 & 3.2k & \underline{39.3} & \textbf{68.7} & \underline{19.2} & \textbf{43.3} & \underline{18.8} & \underline{40.0} \\
\bottomrule
\end{tabular}
\caption{
Performance at a fixed update budget of $\mathcal{B}=51200$ gradient updates, showing configurations up to 16 epochs. All rows within each model use equivalent update budget but vary the epochs-to-samples ratio. For all three models, 16 epochs on 3200 samples substantially outperforms 1 epoch on 51200 samples across all benchmarks.
}
\label{tab:table1}
\end{center}
\end{table*}

We report three metrics: \textbf{Acc@$n$}, the accuracy averaged over $n$ independent generations per problem; \textbf{Pass@$n$}, the fraction of problems solved in at least one of $n$ attempts; and \textbf{Termination}, the fraction of generations that conclude with an end-of-sequence token rather than being truncated. We sample up to 30k tokens per generation to accommodate extended reasoning traces. For AIME we generate 16 responses per problem, while for GPQA, 4 responses due to its larger test set. We use recommended sampling parameters from each model's technical report and vLLM \citep{kwon2023efficient} for efficient inference.

\noindent\textbf{Training.} We load models in bfloat16, use Unsloth optimized kernels \citep{unsloth}, and the 8-bit Adam optimizer \citep{dettmers2022optimizers} with a cosine learning rate schedule. Warmup is set to 10\% of the total update budget for each run. We use a batch size of one, following recent findings that small batch sizes achieve equal or better per-token performance \citep{marek2025small}. We mask the input prompt and compute cross-entropy loss only on response tokens. We conduct a learning rate sweep for each model using 1 epoch on 51200 samples and select the best-performing rate based on benchmark accuracy, then use that learning rate for all subsequent runs. Each configuration is run on a single H100 94GB GPU for up to 24 hours.

\noindent\textbf{Experimental grid.} We train models across dataset sizes from 200 to 51200 samples and epoch counts from 1 to 256, subject to a maximum update budget of 51200. For example, the 200-sample split is trained for up to 256 epochs, while the 51200-sample split is trained for only 1 epoch. Each configuration is trained independently from the base checkpoint with its own warmup and learning rate schedule, rather than evaluating intermediate checkpoints from a single extended run. This design ensures that for any given update budget, we can compare multiple configurations trading off epochs against unique samples.

\subsection{Results}

Figure~\ref{fig1} presents heatmaps of accuracy, pass rate, and termination rate across all combinations of epochs and dataset sizes for Olmo3-7B, averaged over benchmarks. Figure~\ref{fig_heatmaps} presents normalized scores for each benchmark separately, for Olmo3-7B and Qwen3-8B models.
Table~\ref{tab:table1} provides detailed per-benchmark results for all models at a fixed update budget of 51200 gradient updates, showing training runs up to 16 epochs. Figures with all configurations can be found in Appendix \ref{appendix_results}.

We can see a clear and consistent pattern: \textbf{\textcolor{DarkGreen}{fewer unique samples repeated more times yields substantially better performance than training on more data for fewer epochs}}.
For example, at a budget of 51200 updates, Olmo3-7B trained for 32 epochs on 1600 samples reaches an average 39\% accuracy across benchmarks, compared to 17\% for a single epoch on 51200 samples. The same pattern appears across benchmarks and models: on Figure~\ref{fig_heatmaps} all top performances are clearly in the top part of the samples$\times$epochs pyramid. The gains diminish around 32--64 epochs, suggesting a saturation point beyond which additional repetition provides limited benefit. We investigate this saturation in Section~\ref{sec:probing}

\section{Impact of Training Data}

The previous experiments establish the repetition advantage on a general-purpose SFT dataset spanning diverse domains, but whether this phenomenon depends on properties of the training data remains unclear. In this section, we vary data characteristics while keeping the model fixed to Olmo3-7B.

We construct math-focused datasets by distilling long chain-of-thought solutions from various Qwen3 models. We use problems from the NuminaMath-TIR\footnote{\url{https://huggingface.co/datasets/AI-MO/NuminaMath-TIR}} dataset \citep{numina_math_datasets} as prompts and generate solutions using reasoning checkpoints of Qwen3-0.6B and Qwen3-8B as teacher models. We split each distilled dataset into nested subsets from 200 to 25600 samples and train across the same epoch-sample grid as before.

\begin{table*}[t]
\vskip -0.2in
\begin{center}
\small
\begin{tabular}{rrr|rr|rr}
\toprule
\multicolumn{3}{c|}{} & \multicolumn{2}{c|}{Qwen3-0.6B} & \multicolumn{2}{c}{Qwen3-8B} \\
\midrule
Budget & Epochs & Samples & Avg@n & Pass@n & Avg@n & Pass@n \\
\midrule
6.4k & 1 & 6.4k & 2.2 & 13.7 & 10.6 & 36.1 \\
6.4k & 2 & 3.2k & 4.8 & 19.4 & 17.3 & 39.7 \\
6.4k & 4 & 1.6k & 6.9 & 22.3 & 22.1 & 51.0 \\
6.4k & 8 & 800 & 11.4 & 34.1 & \underline{25.4} & \underline{53.6} \\
6.4k & 16 & 400 & \underline{20.6} & \underline{46.5} & \textbf{26.1} & \textbf{55.0} \\
6.4k & 32 & 200 & \textbf{21.6} & \textbf{54.0} & 24.5 & 51.0 \\
\cmidrule(l){1-7}
25.6k & 1 & 25.6k & 3.8 & 16.7 & 13.3 & 36.5 \\
25.6k & 2 & 12.8k & 5.4 & 19.1 & 18.5 & 40.0 \\
25.6k & 4 & 6.4k & 6.5 & 25.9 & 24.1 & 49.9 \\
25.6k & 8 & 3.2k & 10.4 & 29.8 & \underline{33.3} & \underline{64.9} \\
25.6k & 16 & 1.6k & \textbf{20.8} & \underline{49.2} & \textbf{35.5} & \textbf{66.6} \\
25.6k & 32 & 800 & \underline{20.2} & \textbf{49.5} & 30.1 & 63.0 \\
\bottomrule
\end{tabular}
\caption{
Impact of teacher model size on the repetition advantage. Olmo3-7B is trained on math data distilled from Qwen3-0.6B and Qwen3-8B teachers, with results averaged across AIME'24, AIME'25, and GPQA. The repetition advantage persists for both teachers. With the weaker 0.6B teacher, increasing the update budget from 6.4k to 25.6k leads to lower peak performance, echoing the degradation observed in weak-to-strong generalization.
}
\label{tab:teacher}
\end{center}
\end{table*}

\begin{table}[t]
\vskip -0.2in
\setlength{\tabcolsep}{2pt}
\begin{center}
\small
\begin{tabular}{ccrr@{\hskip 0.5cm}rr@{\hskip 0.5cm}rr}
\toprule
Epochs & Subset & \multicolumn{2}{c}{GPQA} & \multicolumn{2}{c}{AIME'24} & \multicolumn{2}{c}{AIME'25} \\
\cmidrule(lr){3-4}
\cmidrule(lr){5-6}
\cmidrule(lr){7-8}
 &  & Acc@4 & Pass@4 & Acc@16 & Pass@16 & Acc@16 & Pass@16 \\
\midrule
1 & Negatives & \textbf{4.5} & \textbf{13.1} & \textbf{15.0} & \textbf{56.7} & {13.3} & {36.7} \\
 & Positives & {3.0} & {9.1} & {13.3} & {33.3} & \textbf{15.0} & \textbf{40.0} \\
\midrule
4 & Negatives & \textbf{11.9} & \textbf{26.8} & \textbf{30.2} & \textbf{73.3} & \textbf{27.5} & \textbf{50.0} \\
 & Positives & {10.4} & {25.3} & {27.3} & {66.7} & {26.0} & {46.7} \\
\midrule
16 & Negatives & \textbf{29.3} & \textbf{54.0} & \textbf{40.0} & \textbf{80.0} & {33.3} & \textbf{66.7} \\
 & Positives & {23.4} & {51.5} & {37.3} & \textbf{80.0} & \textbf{34.2} & {63.3} \\
\bottomrule
\end{tabular}
\caption{
Training on incorrect reasoning traces does not harm performance in long-CoT SFT. Olmo3-7B is trained on positive and negative trajectories distilled from Qwen3-8B, with a fixed update budget of $\mathcal{B}=6.4$k. The repetition advantage holds regardless of trajectory correctness. Surprisingly, training on negatives often matches or exceeds training on positives.
}
\label{tab:neg}
\end{center}
\end{table}

\noindent\textbf{Teacher Model Quality.} Table~\ref{tab:teacher} compares results when training on data distilled from the 0.6B and 8B teachers, for budgets of $\mathcal{B}=25600$ and $\mathcal{B}=6400$. The repetition advantage persists in both settings, with epoch scaling improving performance more reliably than data scaling regardless of teacher size.

The interaction between epochs and data differs between teachers, however. With the smaller 0.6B teacher, the average performance degrades with additional samples; the highest average pass rate for $\mathcal{B}=6400$ is 54.0\%, while for $\mathcal{B}=25600$ it's 49.5\%. This pattern echoes findings in weak-to-strong generalization \citep{w2s}, where student models trained on weaker teacher data can initially exceed teacher performance but degrade with prolonged exposure.

With the larger 8B teacher, the pattern is similar to the previous experiments on the \emph{Dolci SFT 7B} dataset. The model reaches higher absolute performance after sufficient number of epochs, and the performance improves when scaling up the number of data samples. In this case, the highest average pass rate for $\mathcal{B}=6400$ is 55.0\%, while for $\mathcal{B}=25600$ it's 66.6\%. These results suggest that \textbf{\textcolor{DarkGreen}{teacher quality determines whether data scaling remains beneficial, while the repetition advantage itself is robust to teacher choice}}.

\begin{figure*}[t]
  \begin{center}
    \centerline{\includegraphics[width=\textwidth]{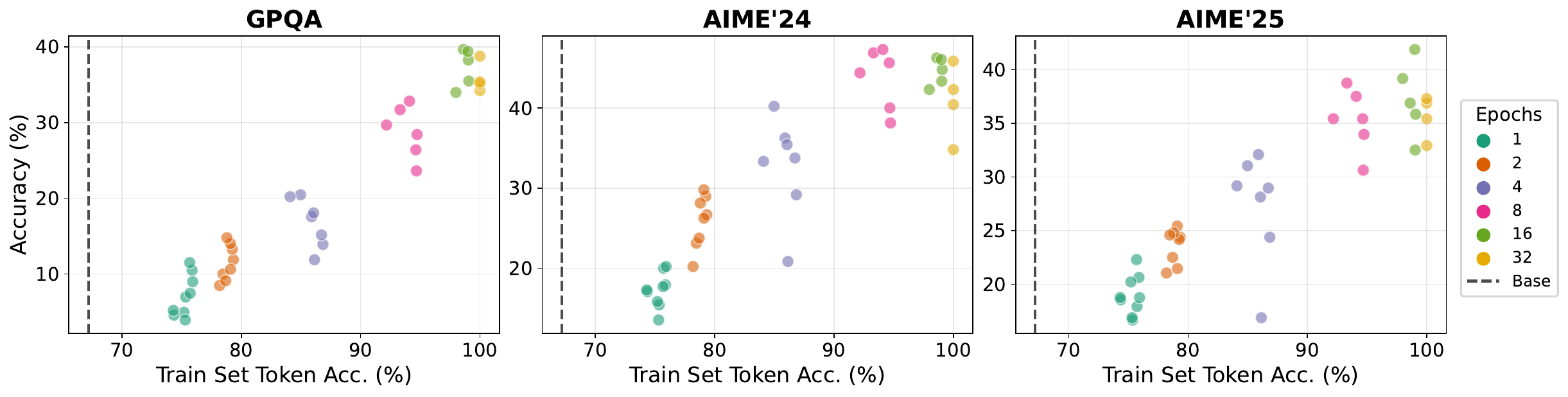}}
    \caption{
Relationship between training set memorization and downstream performance for Olmo3-7B. Points are colored by epoch count; within each epoch group, variation reflects different dataset sizes. Token accuracy on train set increases primarily with epochs, independently of dataset size. Across all benchmarks, performance gains plateau once models approach full memorization, suggesting that token accuracy can serve as a stopping criterion for epoch scaling.
The initial token accuracy of the base model is marked with the vertical line.
    }
    \label{memorization}
  \end{center}
\end{figure*}

\begin{table*}[t]
\vskip -0.2in
\begin{center}
\small
\begin{tabular}{r|rrr|rrr|rrr}
\toprule
\multicolumn{1}{c|}{} & \multicolumn{3}{c|}{Olmo3-7B} & \multicolumn{3}{c|}{Qwen3-8B} & \multicolumn{3}{c}{Qwen3-4B} \\
\midrule
Epochs & Acc@n & Pass@n & T. Acc. & Acc@n & Pass@n & T. Acc. & Acc@n & Pass@n & T. Acc. \\
\midrule
1 & 17.2 & 40.1 & 75.7 & 15.0 & 35.2 & 76.6 & 8.4 & 28.8 & 76.6 \\
2 & 22.6 & 47.5 & 78.8 & 18.9 & 44.1 & 79.4 & 16.2 & 37.4 & 81.5 \\
4 & 27.6 & 54.1 & 84.1 & 25.4 & 54.4 & 84.2 & 21.8 & 43.1 & 88.5 \\
8 & 36.5 & 63.8 & 92.2 & 33.4 & \underline{62.8} & 91.6 & \textbf{29.1} & \underline{50.3} & 96.1 \\
16 & 38.5 & 68.5 & 98.0 & \textbf{37.6} & \textbf{70.9} & 97.7 & \underline{25.7} & \textbf{50.7} & 99.9 \\
32 & \underline{38.8} & \textbf{73.7} & 100.0 & \underline{36.2} & 61.6 & 100.0 & 22.9 & 49.7 & 100.0 \\
64 & \textbf{38.9} & 69.0 & 100.0 & 34.4 & 62.1 & 100.0 & 18.7 & 40.3 & 100.0 \\
128 & 38.4 & \underline{71.5} & 100.0 & 30.4 & 61.4 & 100.0 & 12.4 & 37.6 & 100.0 \\
\bottomrule
\end{tabular}
\caption{
Relationship between training set memorization and average downstream performance at a fixed update budget of $\mathcal{B}=51200$. Token accuracy measures the fraction of response tokens where the model's top prediction matches the training target. Performance improves with epoch count until full memorization, after which gains plateau or degrade.
}
\label{tab:memorization}
\end{center}
\vspace{-3mm}
\end{table*}

\noindent\textbf{Negative Trajectories} If the repetition advantage depends on learning from correct reasoning, training on incorrect traces should degrade performance or exhibit different scaling dynamics. We define \emph{negative trajectories} as chain-of-thought samples where the model's final answer is incorrect.

To test this, we take the data distilled from the Qwen3-8B teacher and partition it by correctness of the final answer. Samples with correct answers form the positive set; those with incorrect answers form the negative set. We construct nested splits from 200 to 6400 samples for each and train Olmo3-7B across the same epoch-sample grid.

From Table~\ref{tab:neg} we find that \textbf{\textcolor{DarkGreen}{training on negative trajectories does not degrade performance}} in long-CoT SFT.
The epoch scaling advantage persists with the same pattern as before. Moreover, the top performance on AIME'24 and GPQA is on-par and even slightly higher when training on negatives than positives, reaching 40.0\% versus 37.3\% on AIME'24 and 29.3\% versus 23.4\% on GPQA. One possible explanation is that negative trajectories come from harder problems where the teacher failed, and exposure to difficult reasoning attempts benefits the student even when the final answer is wrong.

\section{Probing the Repetition Advantage \label{sec:probing}}

Having demonstrated that the repetition advantage is robust across models, benchmarks, and training data sources, we now attempt to understand what drives this phenomenon. We return to the Olmo3-7B models trained on the Dolci dataset from Section~\ref{sec:main} and examine several training dynamics, including memorization, termination behavior, and classical overfitting metrics, searching for signals that might explain why epoch scaling outperforms data scaling. While we identify correlates of improved performance, we do not find a definitive causal mechanism. We present these observations as empirical characterizations that may guide future investigation into the underlying causes.

\noindent\textbf{Memorization signals convergence.} We first investigate training set memorization as a potential indicator of convergence. During SFT, we measure token accuracy on a fixed 200-sample training subset, computing the fraction of response tokens where the model's top next-token prediction matches the target. Figure~\ref{memorization} plots this metric against downstream accuracy for Olmo3-7B. Token accuracy increases primarily with epoch count, independently of dataset size: models trained for 16 epochs achieve near-perfect memorization regardless of whether they see 200 or 3200 unique samples. Across all three benchmarks, \textbf{\textcolor{DarkGreen}{performance improvements plateau once models approach full memorization, which suggests a practical stopping criterion for epoch scaling}}. Table~\ref{tab:memorization} shows this pattern across all three models, revealing that the smaller model memorizes faster and peaks at lower epoch counts.

\noindent\textbf{Termination correlates with performance.} A notable pattern in Figure~\ref{fig1} is the strong correlation between termination rate and accuracy. Single-epoch models terminate only 24\% of generations, while 32-epoch models approach the rate of 89\%. This correlation reflects a causal relationship, where models that fail to terminate cannot produce a final answer, directly limiting their measured accuracy (see Appendix \ref{appendix:termination}).
The increase in termination rate with epoch count suggests that \textbf{\textcolor{DarkGreen}{repeated exposure helps models internalize not just the reasoning patterns but also the structural convention of concluding long reasoning chains}}.
This behavioral convergence appears to require sufficient repetition, as even models trained on 51200 unique samples fail to reliably terminate when trained for only one epoch.

\begin{figure*}[t]
  \vskip -0.2in
  \begin{center}
    \centerline{\includegraphics[width=\textwidth]{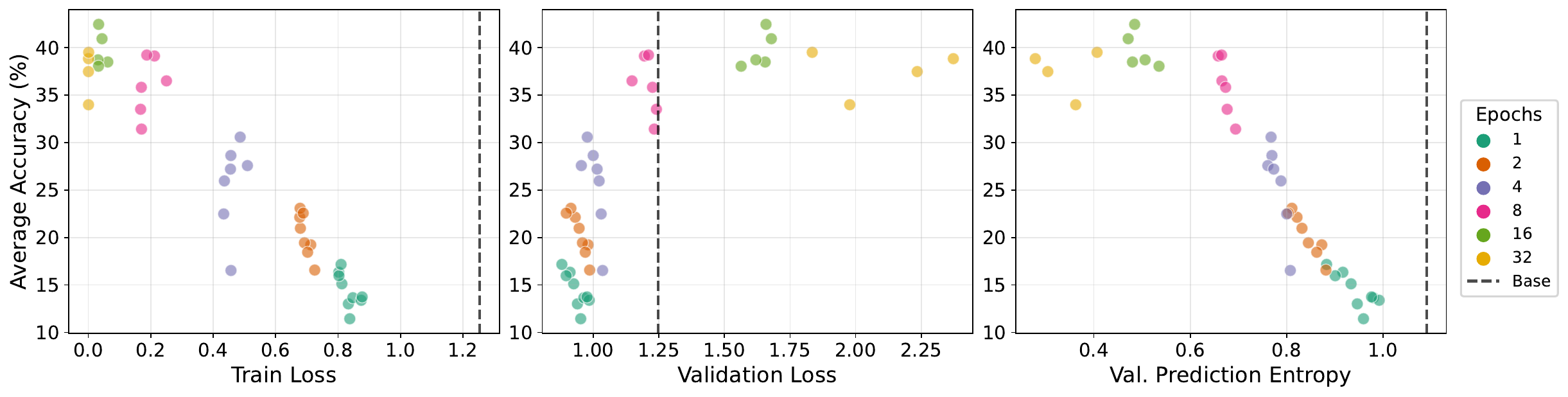}}
    \caption{
    Training dynamics for Olmo3-7B showing the relationship between loss, entropy, and downstream performance averaged over benchmarks.
    As epochs increase, train loss approaches zero while validation loss rises. Prediction entropy also decreases, showing increased model confidence in predictions that diverge from the validation distribution. Despite these indicators, downstream accuracy improves with epoch count. Vertical lines mark base model metrics.
    }
    \label{overfitting}
  \end{center}
\vspace{-2mm}
\end{figure*}

\noindent\textbf{Overfitting paradox.} A natural concern with multi-epoch training is overfitting. Figure~\ref{overfitting} examines this for Olmo3-7B. As epochs increase, train loss approaches zero while validation loss rises substantially. We also measure prediction entropy on the validation set, defined as the average token-level entropy $H = -\sum_i p_i \log p_i$ of the model's output distribution. This metric decreases with epoch count, indicating the model grows more confident in predictions that diverge from the validation distribution. By standard metrics, the model is overfitting. Yet, downstream accuracy improves monotonically with epoch count, suggesting that \textbf{\textcolor{DarkGreen}{validation loss on held-out SFT data is not a reliable metric for reasoning performance}}.

One interpretation is that multi-epoch training elicits latent capabilities already present in the pretrained model, rather than teaching genuinely new skills. The model becomes confident in its own reasoning patterns, which differ from the validation trajectories but nonetheless transfer to held-out benchmarks. This view aligns with recent work on entropy minimization in fine-tuning \citep{agarwal2025entropy} and suggests that SFT may function more as capability elicitation than capability acquisition.

\noindent\textbf{Catastrophic Forgetting} Beyond overfitting, multi-epoch training on small datasets risks catastrophic forgetting, where the model may lose general capabilities while specializing to the narrow training distribution.
To evaluate this, we measure performance on MMLU \citep{hendrycks2021mmlu}, a broad knowledge benchmark spanning 57 subjects. Unlike our reasoning benchmarks, MMLU is evaluated by comparing the model's probability assignments to answer choices rather than generating full responses. We use 5-shot prompting following the standard protocol.

\begin{figure}[t]
  \vskip -0.2in
  \begin{center}
    \centerline{\includegraphics[width=0.7\columnwidth]{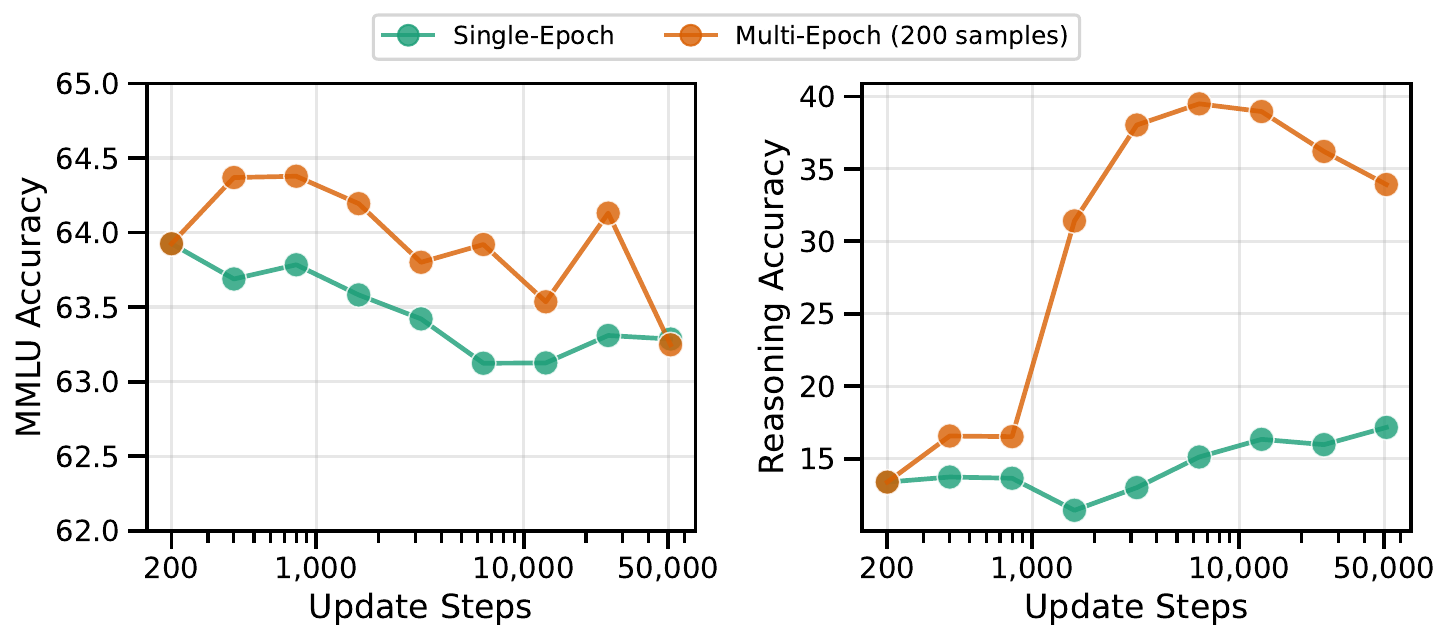}}
    \caption{
    Catastrophic forgetting under epoch scaling versus data scaling for Olmo3-7B. Multi-epoch training on 200 samples is compared against single-epoch training on increasingly large datasets, matched by total update steps. Both approaches exhibit forgetting as measured by MMLU accuracy, with epoch scaling causing \emph{less} degradation. Combined with the large improvement in reasoning accuracy, measured on AIME'24/'25 and GPQA benchmarks, epoch scaling offers a strictly better tradeoff. 
    }
    \label{forgetting}
  \end{center}
\vspace{-2mm}
\end{figure}

Figure~\ref{forgetting} compares two training strategies matched by total gradient updates: scaling epochs on a fixed 200-sample dataset versus scaling dataset size with a single epoch. Both approaches cause some forgetting relative to the base model, as expected when fine-tuning on domain-specific data.
However, \textbf{\textcolor{DarkGreen}{epoch scaling leads to less catastrophic forgetting than data scaling}}.
Combined with the large improvement in reasoning accuracy, epoch scaling offers a strictly better tradeoff.

\section{Related Work}

\noindent\textbf{Data repetition and scaling laws in pretraining.} Scaling laws for language model pretraining characterize how validation loss improves predictably with increased model size, total training tokens, and compute \citep{kaplan2020scalinglaws,hoffmann2022chinchilla}. While these laws are agnostic to whether tokens are unique or repeated, they have commonly been interpreted as motivating the heuristic that, when available, additional fresh data is preferable to revisiting the same corpus.
More directly, recent work studies pretraining in data-constrained regimes where training necessarily becomes multi-epoch. \citet{muennighoff2023dataconstrained} propose data-constrained scaling laws that explicitly model the \emph{decreasing marginal value of repeated tokens}, and empirically find that repeating a fixed corpus for a small number of epochs (on the order of a few passes) can be nearly as effective for loss as training on equivalently-sized fresh tokens, while the returns from further repetition decay sharply. Relatedly, recent work on diffusion language models shows that, in data-constrained \emph{pretraining} regimes, extensive data repetition can be beneficial, with diffusion objectives extracting substantially more value per unique token than autoregressive training \citep{ni2025superdatalearner}.
Our work contrasts with this pretraining-focused literature by showing that the ``avoid repetition'' heuristic does not transfer to supervised fine-tuning on long chain-of-thought data; on the contrary, repetition substantially improves convergence and downstream performance.

\noindent\textbf{Multi-epoch SFT in post-training practice.} Although single-pass training is often treated as the default in instruction tuning, many recent training pipelines perform supervised fine-tuning for multiple epochs as part of post-training, often without isolating epoch count as a studied variable.
Examples include: 1) Olmo 3 reports training on SFT data, consisting of over 2M samples, for two epochs \citep{teamolmo2025olmo3}.
2) DeepSeek-R1 similarly includes an SFT phase that fine-tunes its base model for 2-3 epochs on a large curated set prior to reinforcement learning \citep{guo2025deepseekr1}.
3) Llama-3 trains SFT for ``multiple epochs''~\citep{llama3}.
4) LIMO trains for 15 epochs on a curated reasoning set \citep{ye2025limo}, while 5) \citet{muennighoff2025s1simpletesttimescaling} train a model on long-CoT data for 5 epochs.
Across these releases, epoch counts are typically presented as recipe details rather than as ablated design choices.
Our work provides a controlled, compute-matched comparison of \emph{epoch scaling} versus \emph{unique-data scaling} in long-CoT SFT, showing that multi-epoch training can be a better strategy even when additional training tokens are available.

\noindent\textbf{Memorization, overfitting, and training dynamics.} Classic results in deep learning challenge the view that memorization necessarily harms generalization.
\citet{arpit2017memorization} show that deep networks tend to learn simple patterns before memorizing noise.
For language modeling specifically, \citet{tirumala2022memorizationwo} study \emph{exact memorization} throughout training and characterize how memorization depends on model size, dataset size, and optimization choices.
Complementing these empirical findings, \citet{feldman2019longtail} provides a theoretical perspective arguing that memorization can be necessary for generalization on long-tailed data distributions.
We connect to this literature by showing that, in long-CoT supervised fine-tuning, downstream gains from repetition saturate when the model reaches near-perfect token-level accuracy on the training demonstrations.

\section{Conclusion}
We show that supervised fine-tuning on long chain-of-thought data can defy standard machine learning intuition. Under a fixed update budget, training for more epochs on smaller datasets substantially outperforms training on larger datasets, and this repetition advantage holds across models, benchmarks, and training data sources studied in this work. Despite its robustness, the mechanism underlying the repetition advantage remains poorly understood. While training token accuracy provides a practical stopping signal for epoch scaling, the optimal dataset size is data- and model-dependent, and principled criteria for selecting it a priori remain elusive. 
We argue that explaining why memorization under repetition improves generalization in reasoning SFT is an important open problem. More broadly, our results suggest that both epoch count and dataset size should be treated as first-class decision variables in reasoning SFT, rather than defaulting to single-epoch training on the largest available dataset.

\subsection*{Acknowledgements}

The authors gratefully acknowledge the HPC resources provided by the Erlangen National High Performance Computing Center (NHR@FAU) of the Friedrich-Alexander-Universität Erlangen-Nürnberg (FAU) under the BayernKI project `v115be`. BayernKI funding is provided by Bavarian state authorities.

\bibliography{colm2026_conference}
\bibliographystyle{colm2026_conference}

\newpage
\appendix

\section{Hyperparameters}

\begin{table}[H]
\centering
\begin{tabular}{ll}
\hline
\textbf{Hyperparameter} & \textbf{Value} \\
\hline
GPUs & 1 H100 94GB per run \\
Optimizer & 8-bit Adam \\
Weight Decay & 0.0 \\
Beta1 & 0.9 \\
Beta2 & 0.999 \\
Grad. Norm Clipping & 1.0 \\
Batch Size & 1 \\
LR Scheduler & Cosine \\
Warmup Steps & 10\% of all updates \\
RNG Seed & 42 \\
\hline
\end{tabular}
\end{table}

\begin{table}[H]
\centering
\begin{tabular}{l|ccc}
\hline
& \textbf{Olmo3 7B} & \textbf{Qwen3 8B} & \textbf{Qwen3 4B} \\
\hline
Learning Rate & 2e-5 & 2e-5 & 3e-5 \\
\hline
\end{tabular}
\end{table}

\section{Sampling Parameters}

\begin{table}[H]
\centering
\label{tab:sampling-parameters}
\begin{tabular}{lccc}
\toprule
Model & Temperature & \texttt{top\_k} & \texttt{top\_p} \\
\midrule
Qwen3 & 0.6 & 20 & 0.95 \\
Olmo3 & 0.6 & 50 & 0.95 \\
\bottomrule
\end{tabular}
\end{table}

\section{Termination-conditioned Results}
\label{appendix:termination}

\begin{table}[H]
    \centering
    \caption{
        Full GPQA evaluation results for Olmo3-7B at a fixed update budget
        of $\mathcal{B}=51{,}200$. Evaluation uses 198 problems with four
        generations per problem. ``Correct
        (\% terminated)'' reports accuracy conditioned on the generation
        reaching an end-of-sequence token.
    }
    \label{tab:olmo3_gpqa_termination}
    \small
    \setlength{\tabcolsep}{5pt}
    \begin{tabular}{@{}rr rr rr rr r@{}}
        \toprule
        \multicolumn{2}{c}{Training}
        & \multicolumn{2}{c}{Terminated}
        & \multicolumn{2}{c}{Outcome count}
        & \multicolumn{2}{c}{Correct}
        & \multicolumn{1}{c}{Terminated \& wrong} \\
        \cmidrule(lr){1-2}
        \cmidrule(lr){3-4}
        \cmidrule(lr){5-6}
        \cmidrule(lr){7-8}
        \cmidrule(l){9-9}
        Epochs
        & Samples
        & $n$
        & \% total
        & Correct
        & Wrong
        & \% total
        & \% terminated
        & $n$ \\
        \midrule
        1  & 51{,}200 & 244          & 30.81          & 105          & 687          & 13.26          & 43.03          & \textbf{139} \\
        4  & 12{,}800 & 488          & 61.62          & 223          & 569          & 28.16          & \textbf{45.70} & 265          \\
        16 & 3{,}200  & \textbf{781} & \textbf{98.61} & \textbf{317} & \textbf{475} & \textbf{40.03} & 40.59          & 464          \\
        \bottomrule
    \end{tabular}
\end{table}

\section{Training Loss.}

\begin{figure}[H]
    \centering
    \includegraphics[width=\textwidth]{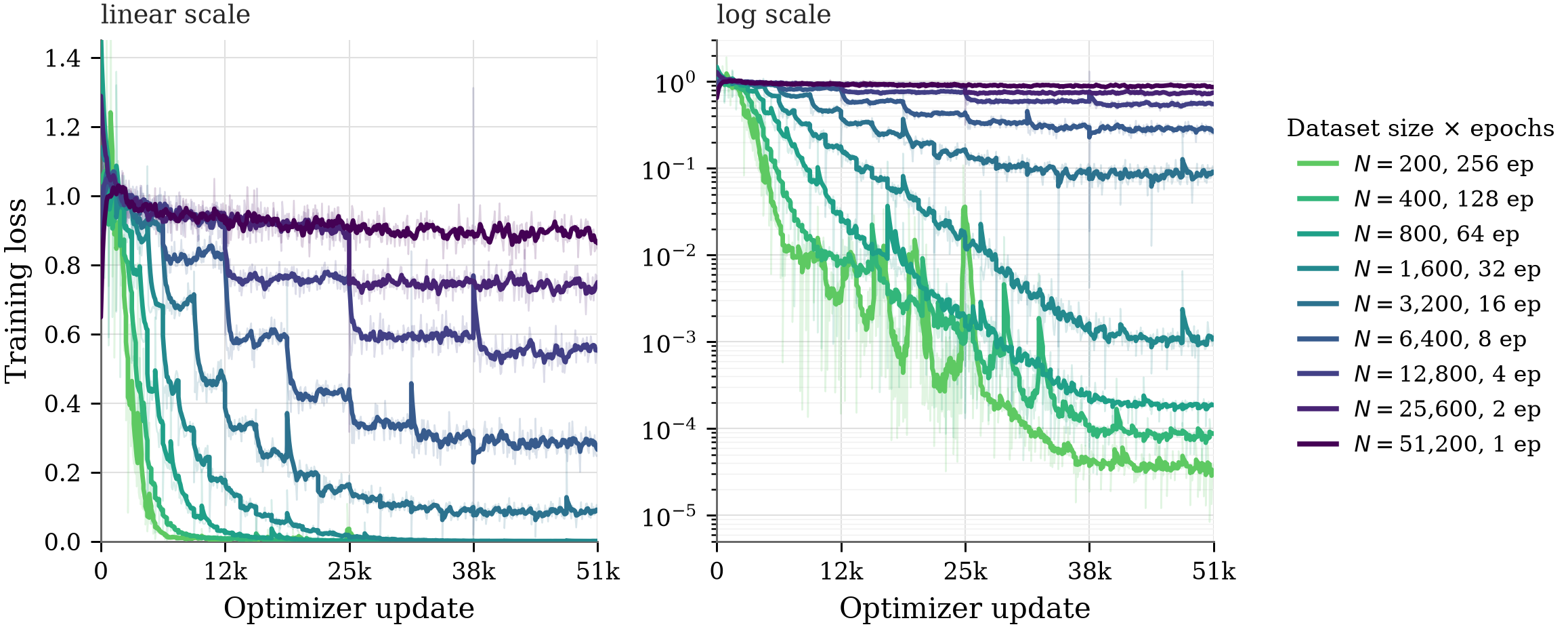}
    \caption{
        Training-loss curves for Olmo3-7B configurations with a fixed update
        budget of $\mathcal{B}=51{,}200$, shown on linear (left) and logarithmic
        (right) scales. Each configuration varies the number of unique training
        samples and epochs while keeping the update budget fixed.
    }
    \label{fig:loss_curves_51200}
\end{figure}

\section{Full Results.}
\label{appendix_results}

\subsection{Dolci Dataset}

\begin{figure}[H]
  \begin{center}
    \centerline{\includegraphics[width=0.7\textwidth]{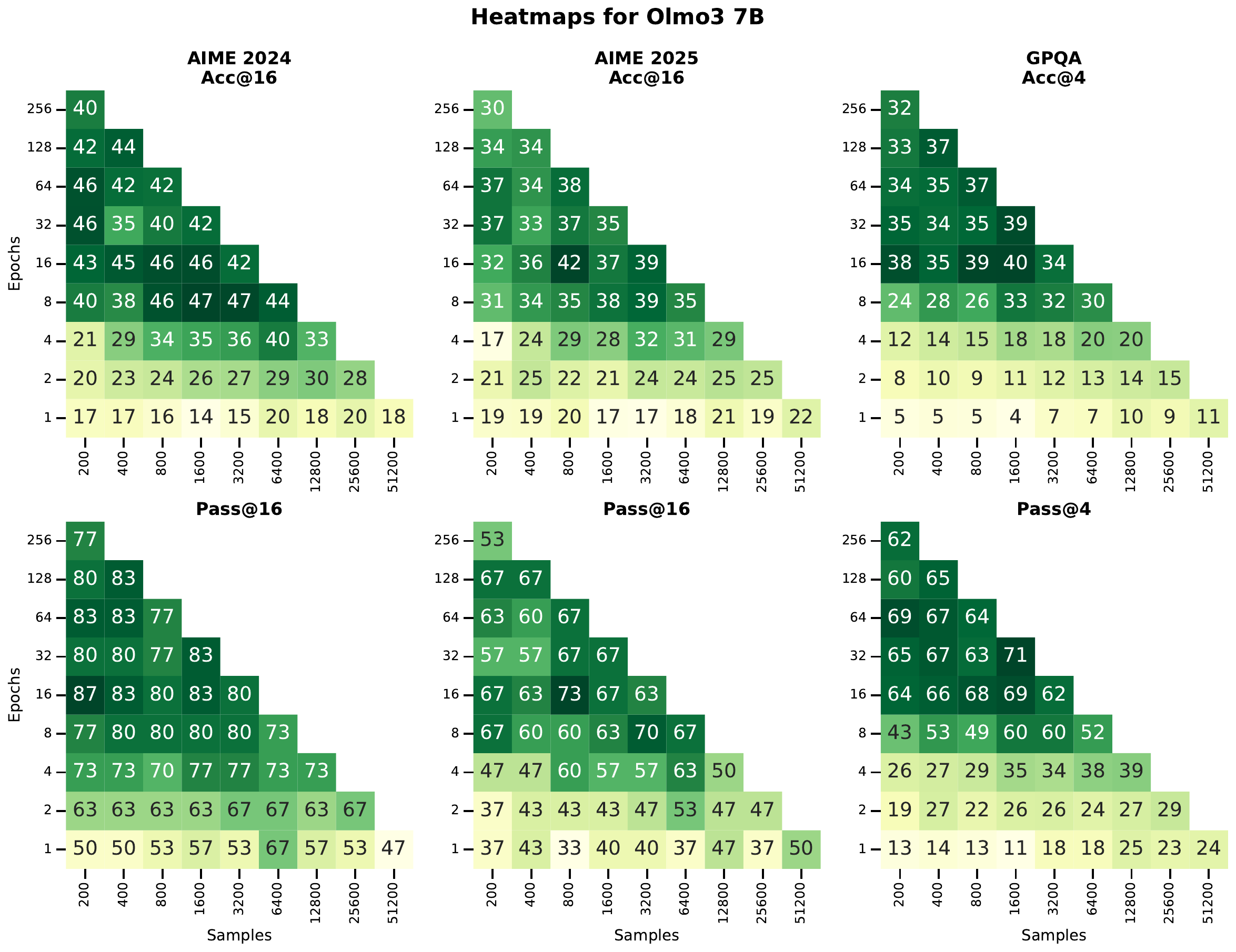}}
        \caption{Dolci dataset results for Olmo3-7B.    \label{appendix_dolci_olmo}}
  \end{center}
\end{figure}

\begin{figure}[H]
  \begin{center}
    \centerline{\includegraphics[width=0.7\textwidth]{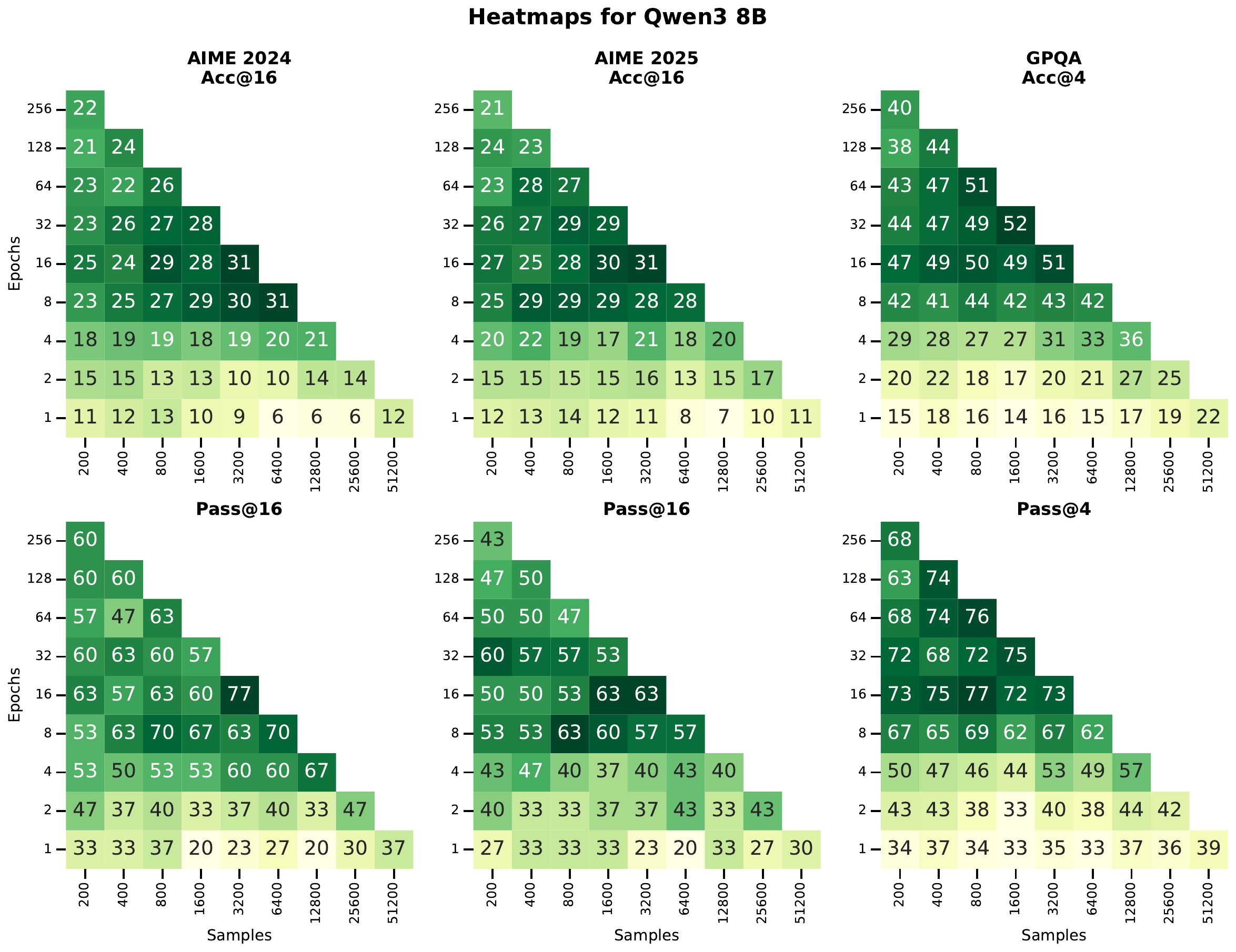}}
        \caption{Dolci dataset results for Qwen3-8B.}
    \label{appendix_dolci_qwen8b}
  \end{center}
\end{figure}

\begin{figure}[H]
  \begin{center}
    \centerline{\includegraphics[width=0.7\textwidth]{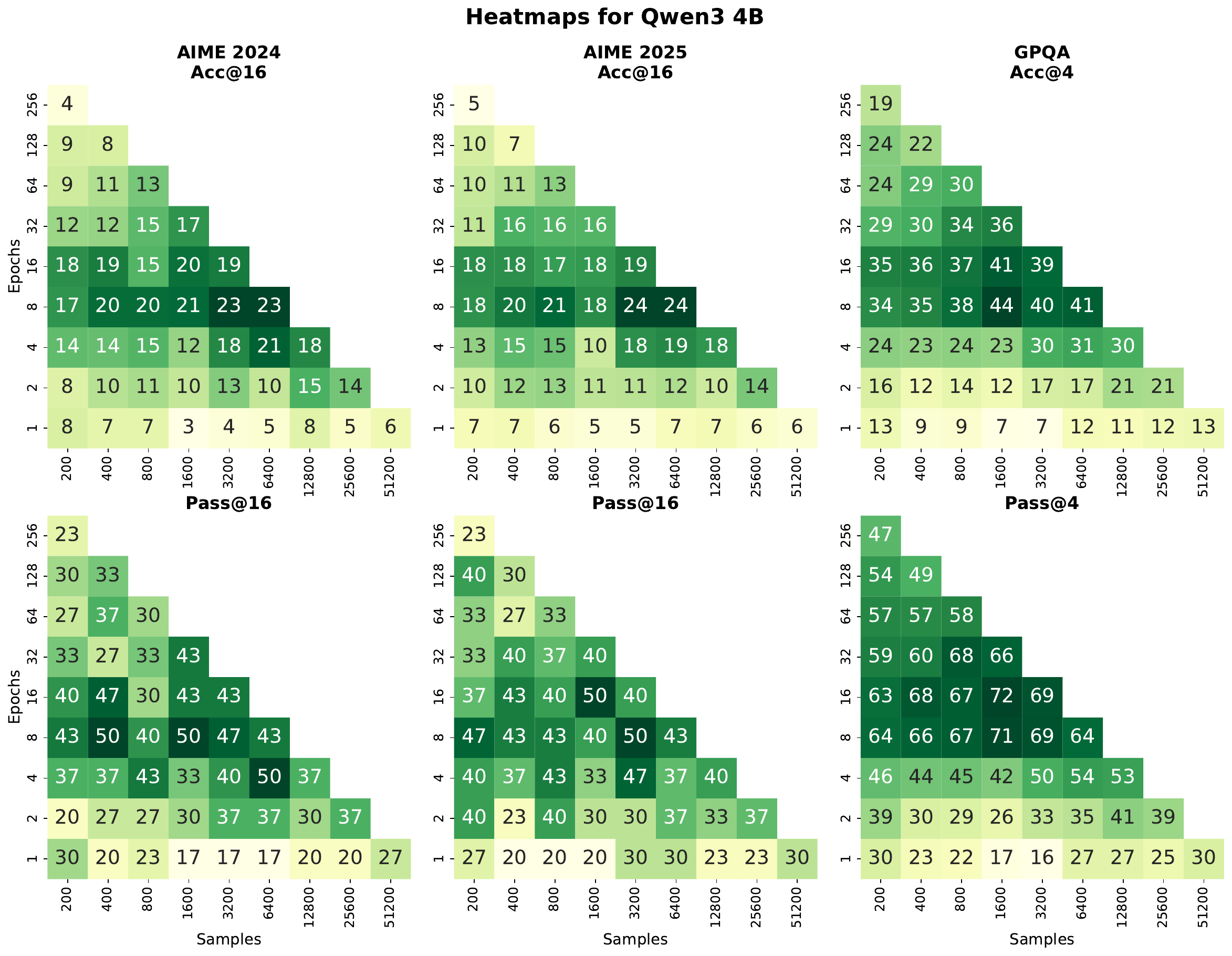}}
        \caption{Dolci dataset results for Qwen3-4B.     \label{appendix_dolci_qwen4b}}
  \end{center}
\end{figure}

Figures~\ref{appendix_dolci_olmo}--\ref{appendix_dolci_qwen4b} show results on the Dolci dataset for three model backbones. Across all models, training for more epochs on smaller datasets consistently outperforms training on larger datasets for fewer epochs.

\subsection{Qwen3 Distills}
\begin{figure}[H]
  \begin{center}
    \centerline{\includegraphics[width=0.7\textwidth]{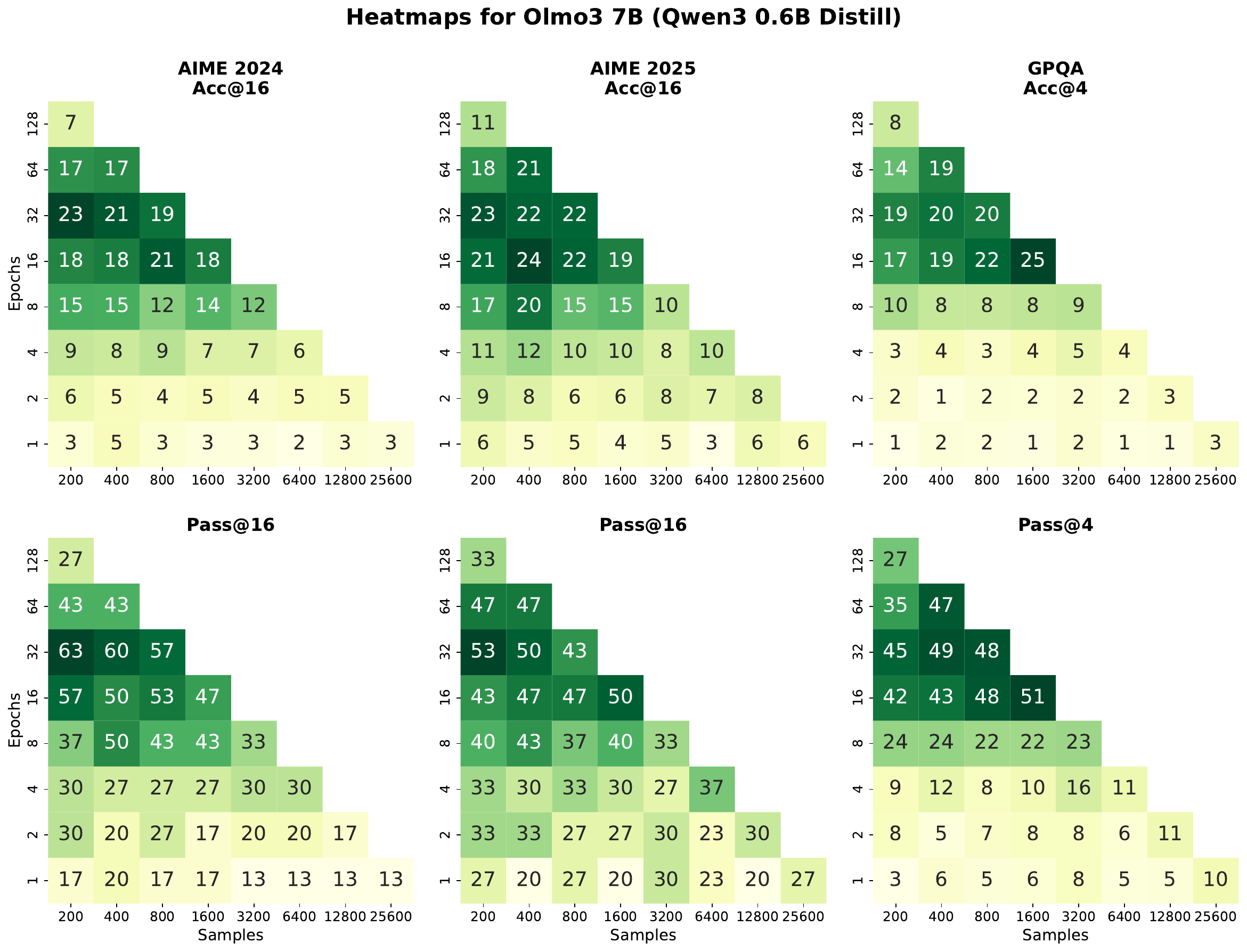}}
        \caption{Results for distillation from a Qwen3-0.6B teacher. Despite weaker teacher signals, repetition continues to improve downstream accuracy.    \label{appendix_06b}}
  \end{center}
\end{figure}

\begin{figure}[H]
  \begin{center}
    \centerline{\includegraphics[width=0.7\textwidth]{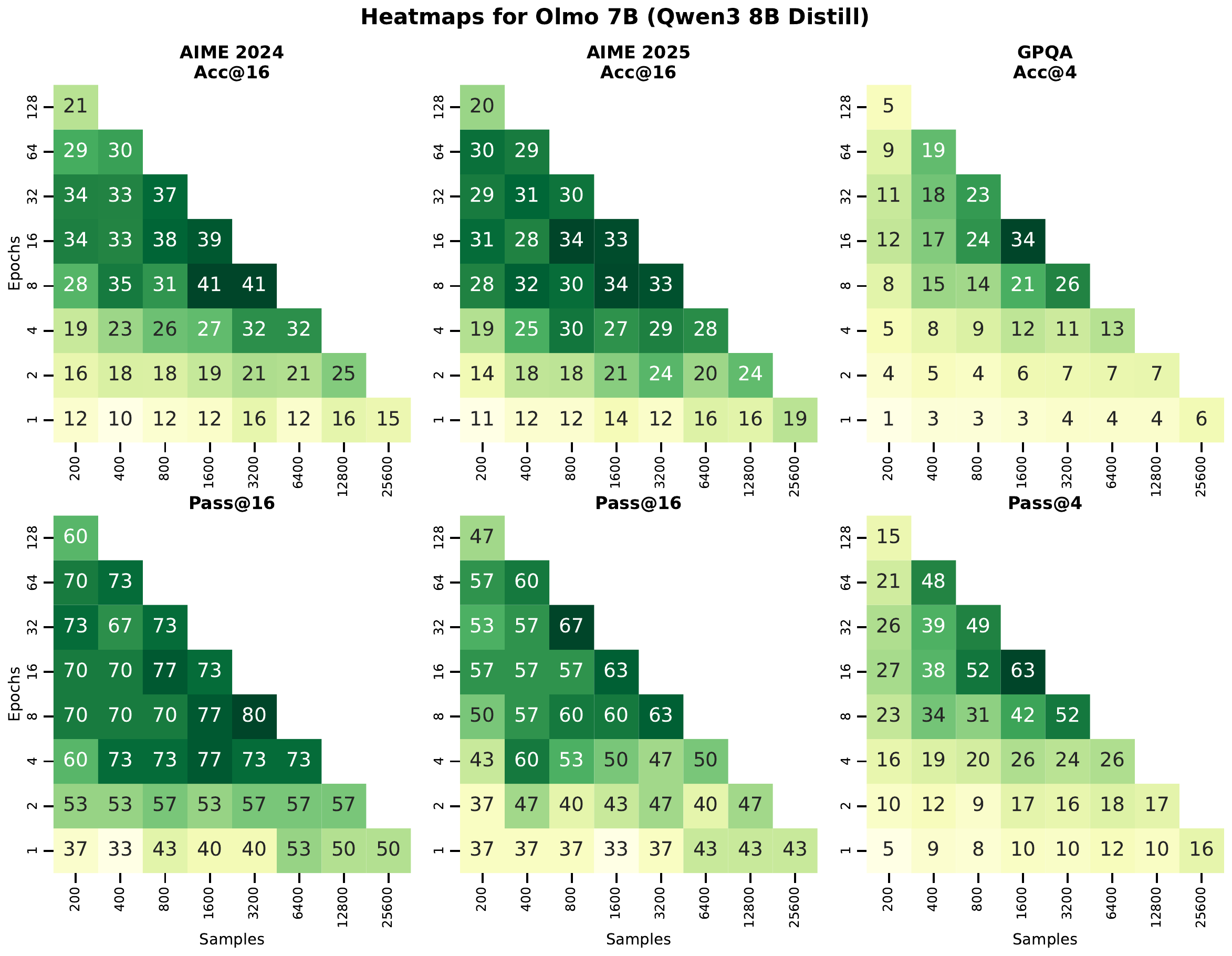}}
            \caption{Results for distillation from a Qwen3-8B teacher. Stronger teachers increase overall performance but do not eliminate the repetition advantage.     \label{appendix_8b}}
  \end{center}
\end{figure}

Figures~\ref{appendix_06b} and~\ref{appendix_8b} examine the effect of teacher model size in distillation. While stronger teachers improve absolute performance, the repetition advantage remains robust: multi-epoch training on smaller distilled datasets consistently outperforms scaling unique samples.

\subsection{Qwen3 8B Distill; Pos. vs Neg.}
\begin{figure}[H]
  \begin{center}
    \centerline{\includegraphics[width=0.7\textwidth]{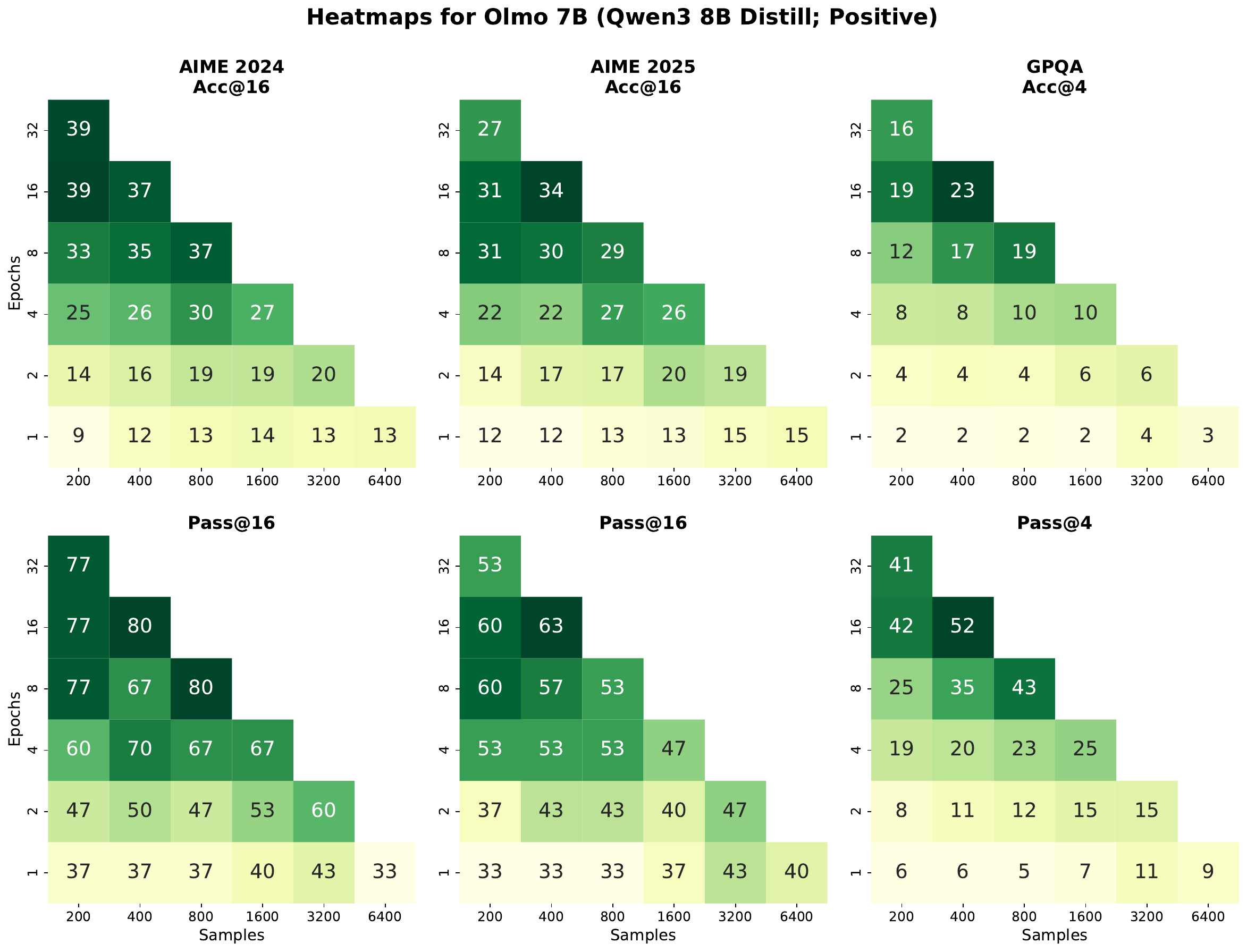}}
    \caption{Results using only correct (positive) distilled samples from a Qwen3-8B teacher.
    \label{appendix_8b_pos}}
  \end{center}
\end{figure}

\begin{figure}[H]
  \begin{center}
    \centerline{\includegraphics[width=0.7\textwidth]{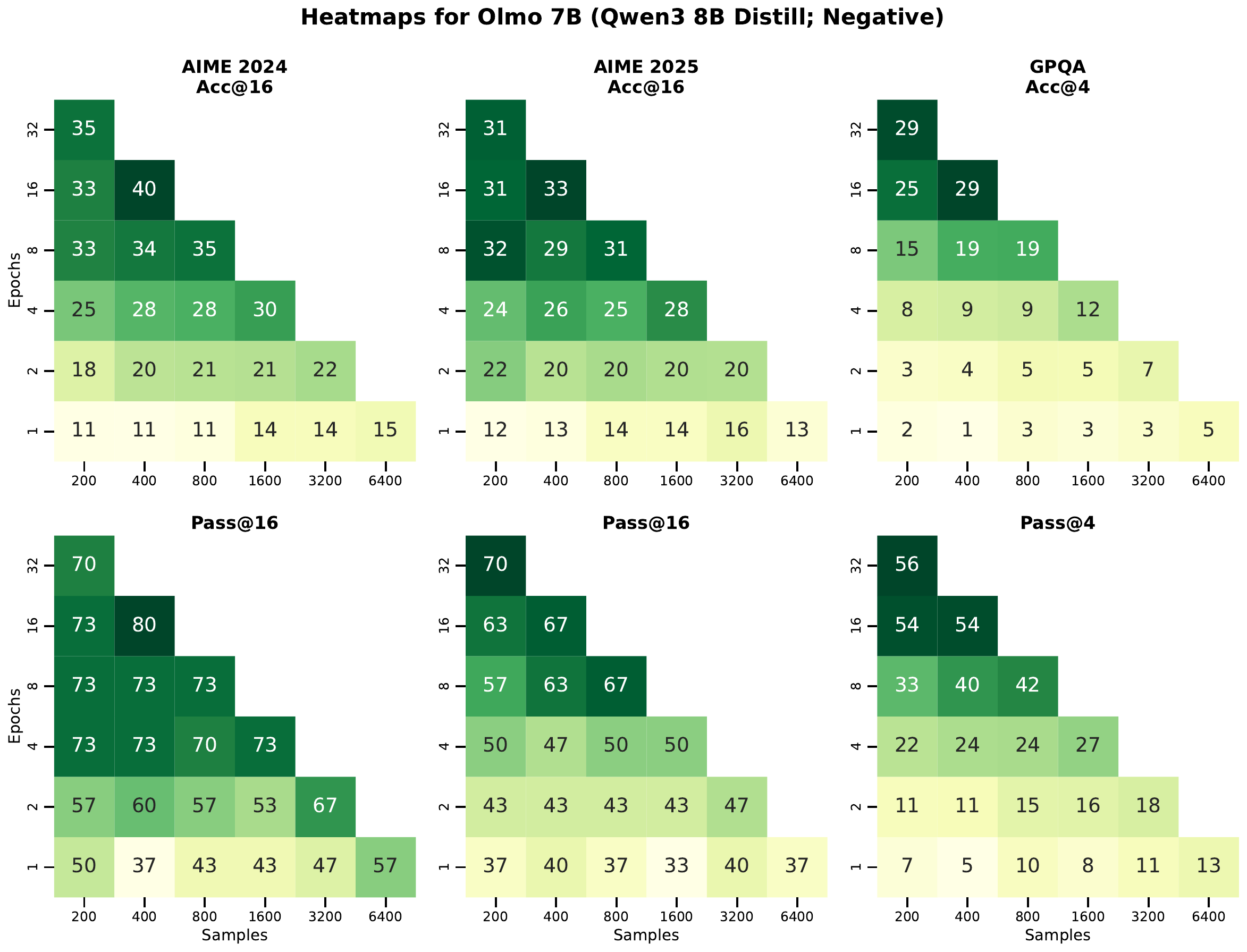}}
        \caption{Results using incorrect (negative) distilled samples from a Qwen3-8B teacher.
        \label{appendix_8b_neg}}
  \end{center}
\end{figure}

Figures~\ref{appendix_8b_pos} and~\ref{appendix_8b_neg} separate distilled samples by correctness.

\end{document}